\documentclass[runningheads]{llncs}
\usepackage[T1]{fontenc}
\usepackage{graphicx}
\usepackage{colortbl}
\usepackage{standalone}
\usepackage{tabularray}
\usepackage{multirow}
\usepackage{tikz}
\usepackage{pgfplots}
\pgfplotsset{
    table/search path={figures},
}
\pgfplotsset{compat=1.18}

\usepackage{bm}
\renewcommand{\vec}[1]{\bm{#1}}
\newcommand{\T}{T}
\providecommand{\mat}[1]{\mathsf{#1}}
\usepackage{mathtools}

\DeclarePairedDelimiterX{\norm}[1]{\lVert}{\rVert}{#1}
\DeclarePairedDelimiterX{\abs}[1]{\lvert}{\rvert}{#1}

\usepackage{bbm}
\newcommand{\C}{\mathbbm{C}}
\newcommand{\R}{\mathbbm{R}}

\DeclareMathOperator{\vectorize}{vec}
\usepackage{gensymb}  

\newcommand{\x}{\vec{x}}
\newcommand{\A}{\mat{A}}
\newcommand{\HH}{\mat{H}}
\newcommand{\bH}{\HH}
\newcommand{\bR}{\mat{R}}
\newcommand{\bt}{\vec{t}}
\newcommand{\bK}{\mat{K}}
\newcommand{\bn}{\vec{n}}

\usepackage{eccvabbrv}

\usepackage{graphicx}
\usepackage{booktabs}

\usepackage[detect-all,per-mode=symbol]{siunitx}

\usepackage{standalone}
\usepackage{tikz}
\usetikzlibrary{calc}
\usepackage{pgfplots}
\pgfplotsset{
    table/search path={plots/noise,plots/numerical_stability,plots/ransac,plots/hpatches_time_maa},
}
\pgfplotsset{
    fitzgibbon_one_sided/.style={
        mark=none, red, thick, solid
    },
    nakano_one_sided/.style={
        mark=none, blue, thick, solid
    },
    wadenback_one_sided/.style={
        mark=none, green!80!black, thick
    },
    affine_dist/.style={
        mark=none, yellow!80!black, thick, dashed
    },
    ori/.style={
        mark=none, magenta!80!black, thick, dotted
    },
} 
\pgfplotsset{compat=1.18}

\usepackage[accsupp]{axessibility}  

\usepackage{hyperref}

\usepackage[capitalise]{cleveref}

\usepackage{orcidlink}
\usepackage{bbding} 

\begin{document}

\title{Radial distortion homography estimation from affine-covariant or orientation-covariant features} 

\titlerunning{Radial distortion homography estimation}

\author{Marcus Valtonen \"{O}rnhag\Envelope\orcidlink{0000-0001-8687-227X} \and Stefan Adalbj\"{o}rnsson\orcidlink{0000-0003-0403-1544}}

\authorrunning{M. Valtonen \"{O}rnhag and S. Adalbj\"{o}rnsson}

\institute{Ericsson Research, Sweden\\
\email{\{marcus.valtonen.ornhag,stefan.adalbjornsson\}@ericsson.com}}

\maketitle

\begin{abstract}

We present minimal solvers for joint homography and radial distortion estimation using affine- or orientation-covariant features.
The core contribution is a set of novel constraints that exploit local feature geometry inherent to affine descriptors, enabling solvers that require fewer correspondences than point-based methods.
Using modern robust estimation frameworks, we demonstrate that the proposed solvers are advantageous to the
current state-of-the-art with a significant speed-up in the presence of outliers while maintaining comparable or superior estimation accuracy. The proposed solvers are tested on both
synthetic and real images containing challenging fisheye distortions.\footnote{Code available here: \url{https://github.com/marcusvaltonen/HomLib}.}

\end{abstract}

\section{Introduction}
\label{sec:intro}

For more than twenty years~\cite{matas-etal-icpr-2002} affine features have been used in geometric computer vision applications.
Early work includes~\cite{perdoch-etal-icpr-2006}, where it was shown how to estimate the epipolar geometry of an
uncalibrated pair of cameras from two correspondences of local affine frames (LAFs). Since then, there have been numerous methods that exploit their intrinsically encoded properties. 
Recent work include homography estimation~\cite{barath-hajder-2017},
relative pose estimation~\cite{barath-hajder-2018,eichhardt-chetverikov-eccv-2018,guan2022relative},
including partially calibrated cameras~\cite{hruby-etal-eccv-2024} and
generalized relative pose~\cite{guan2023minimal}, as well as the
absolute pose problem~\cite{Ventura_2023_ICCV}.

The main theoretical advantage of leveraging affine features in epipolar geometry is the smaller sample size needed in the resulting solver, thus requiring fewer RANSAC iterations in order to to draw an all-inlier subset. The main criticism against using affine solvers has been that they have not been working well in practice; however, Barath~\etal{}~\cite{barath-etal-eccv-2020}
provided a framework for improving affine solvers for various camera geometry estimation pipelines, making them robust and reliable for real-world data. Another issue,
raised in~\eg{},~\cite{valtonen-ornhag-jaenal-eccv-2024}, is the added computational time for extracting the affine features themselves.
To avoid this computational overhead,
Barath and Kukelova~\cite{barath-kukelova-2019-iccv} showed how scale- and orientation-covariant features can be interpreted
as local affine features, utilizing the intrinsic geometric properties of the feature descriptors, thus providing more local information than the point correspondences alone, while not having to compute the affine features explicitly.
They derived a set of constraints characterizing the relationship between affine feature descriptors and the scale- and orientation-covariant counterpart, from which they constructed a
homography solver requiring only two scale- and orientation-covariant features. Ding~\etal{} showed how to incorporate a known
reference direction~\cite{ding-etal-accv-2020} from an IMU using these constraints and further assumptions on the plane normals (\eg{}, vertical and horizontal walls).
In~\cite{barath-kukelova-2022-eccv}, scale- and orientation-covariant features were used for general problems in epipolar
geometry. This was later used for the application of autonomous driving in~\cite{valtonen-ornhag-jaenal-eccv-2024}, where upright essential
matrices were estimated with great computational improvement compared to point-based and affine-based methods. Ventura~\etal{}~\cite{Ventura_2024_CVPR}
derived new constraints for the absolute pose problem from two oriented features and the upright case (using IMU data) from a single scale- and orientation-covariant feature.
Barath~\cite{barath-2023} further showed how to leverage any affine-based solver by approximating affine features directly from
scale- and orientation-covariant features.

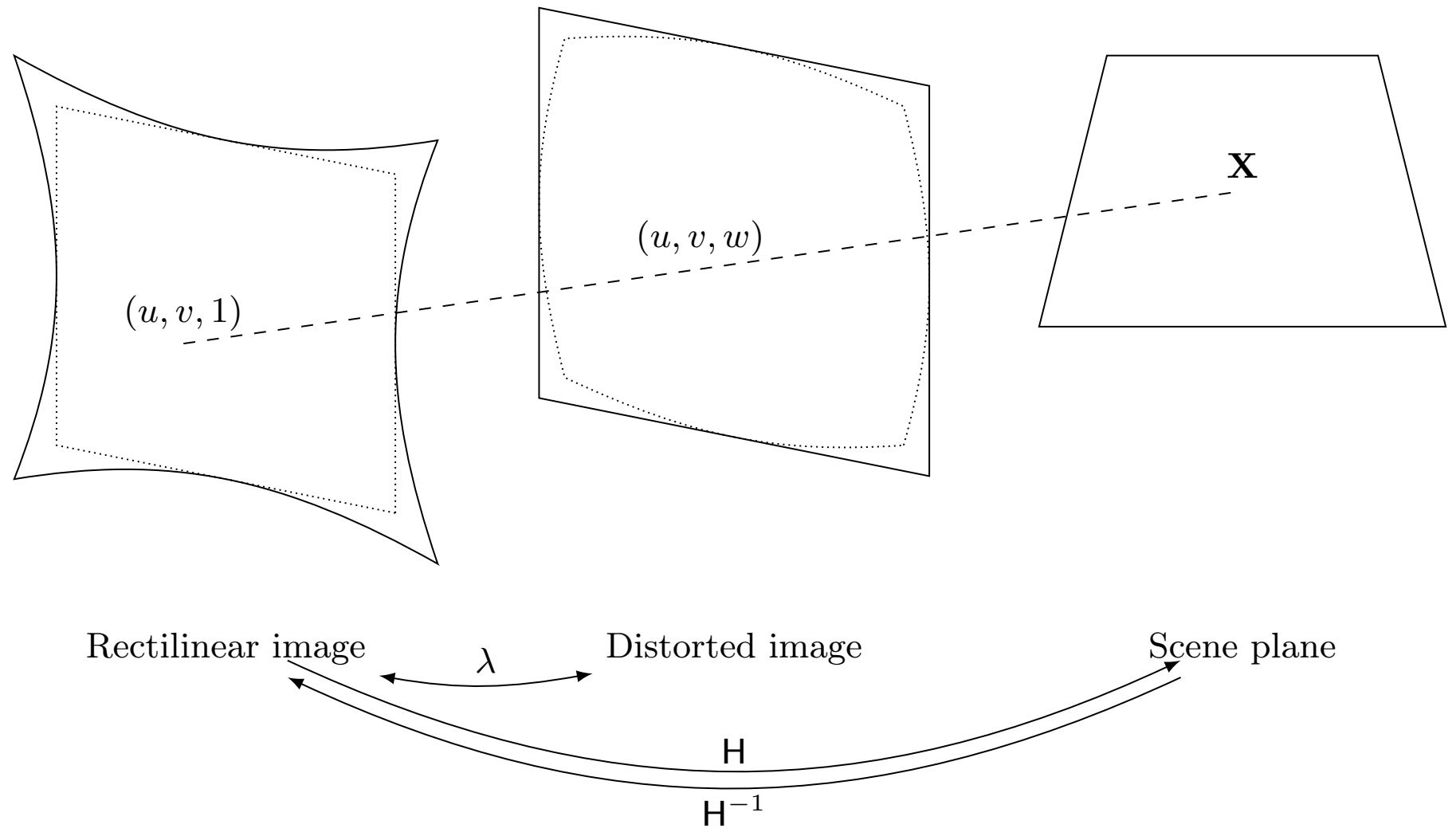
\begin{figure}[t!]
    \centering
    \begin{tikzpicture}[scale=1.5]
    \begin{scope}[shift={(-3,0)}]
        \node[align=center] (A) at (0,-2) {Distorted image};
        \begin{scope}[yslant=-0.2,shift={(0,0.4)}]
            \coordinate (p1) at (1,-1);
            \coordinate (p2) at (1,1);
            \coordinate (p3) at (-1,1);
            \coordinate (p4) at (-1,-1);
            \coordinate (AC) at (-0.2,-0.2);
            \pgfmathsetmacro{\tmpang}{15}
            \draw[densely dotted] (p1) to[in=-90+\tmpang,out=90-\tmpang] coordinate[midway] (mid12) (p2) to[in=\tmpang,out=180-\tmpang] coordinate[midway] (mid23) (p3) to[in=90+\tmpang,out=-90-\tmpang] coordinate[midway] (mid34) (p4) to[in=-180+\tmpang,out=-\tmpang] coordinate[midway] (mid41) cycle;
            \draw ($(mid12)+(mid41)$) -- ($(mid12)+(mid23)$) -- ($(mid23)+(mid34)$) -- ($(mid34)+(mid41)$) -- cycle;
        \end{scope}
    \end{scope}
    \begin{scope}[shift={(-6,0)}]
        \node[align=center] (B) at (0,-2) {Rectilinear image};
        \begin{scope}[yslant=-0.2,scale=1.25]
            \coordinate (p1) at (1,-1);
            \coordinate (p2) at (1,1);
            \coordinate (p3) at (-1,1);
            \coordinate (p4) at (-1,-1);
            \coordinate (BC) at (-0.2,-0.2);
            \pgfmathsetmacro{\tmpang}{-20}
            \draw (p1) to[in=-90+\tmpang,out=90-\tmpang] coordinate[midway] (mid12) (p2) to[in=\tmpang,out=180-\tmpang] coordinate[midway] (mid23) (p3) to[in=90+\tmpang,out=-90-\tmpang] coordinate[midway] (mid34) (p4) to[in=-180+\tmpang,out=-\tmpang] coordinate[midway] (mid41) cycle;
            \draw[densely dotted] ($(mid12)+(mid41)$) -- ($(mid12)+(mid23)$) -- ($(mid23)+(mid34)$) -- ($(mid34)+(mid41)$) -- cycle;
        \end{scope}
    \end{scope}

    \begin{scope}
        \node[align=center] (C) at (0,-2) {Scene plane};
        \begin{scope}[shift={(0,0.5)}]
            \coordinate (p1) at (1.2,-0.6);
            \coordinate (p2) at (0.8,1);
            \coordinate (p3) at (-0.8,1);
            \coordinate (p4) at (-1.2,-0.6);
            \coordinate (CC) at (0,0.2);
            \draw (p1) -- (p2) -- (p3) -- (p4) -- cycle;
        \end{scope}
    \end{scope}

    \draw[latex-latex] (A) to[out=-170,in=-10] node[above,midway] {$\lambda$} (B);
    \draw[latex-] (B) to[out=-25,in=205] node[above,midway,yshift=1mm] {$\mat{H}$} (C);
    \begin{scope}[transform canvas={shift={(0,0.15)}}]
        \draw[-latex] (B) to[out=-25,in=205] node[below,midway,yshift=-1mm] {$\mat{H}^{-1}$} (C);
    \end{scope}

    \draw[dashed] (BC) -- (AC);
    \draw[dashed] (AC) -- (CC);

    \node[above] at (AC) {$(u,v,w)$};
    \node[above] at (BC) {$(u,v,1)$};
    \node[above] at (CC) {$\mathbf{X}$};
\end{tikzpicture}
    \caption{The one-sided radial distortion homography~$\mat{H}$ is a mapping between rectified images. The unknown distortion coefficient~$\lambda$ is applied to one of the images.}
    \label{fig:firstpage}
\end{figure}

Homography estimation is at the core of many computer vision applications, \eg{}, the famous camera calibration method
implemented in OpenCV and many other frameworks, known as Zhang's method~\cite{zhang}, where a planar calibration target with a
checkerboard pattern is placed at different positions in the field-of-view of the camera. Even in the era of deep learning,
homography estimation has not become obsolete, but instead been incorporated as geometric priors~\cite{le-etal-cvpr-2020,cao-etal-cvpr-2023,Ye_2021_ICCV}.
Due to the computational efficiency of state-of-the-art methods for homography estimation, it is integrated in many
modern augmented reality (AR) and mixed reality (MR) platforms, and plays a crucial part for localizing planar surfaces on such devices. Early examples of how to 
place augmented objects on and subsequently interact with them include~\cite{simon-etal-2000,zhou-etal-2012}.
Recently, Nakano~\cite{nakano-icpr-2025} argued that the one-sided radial distortion homography estimation case is more interesting
for AR/MR applications as one image is often anchored in 3D space thus unaffected (or already corrected for) lens distortion artifacts, see~\cref{fig:firstpage}. Another common scenario
for this setup is in cases of relocalization in an existing environment, where the 3D reconstruction has been corrected for distortion artifacts, but the current intrinsic parameters of the devices are unknown. Failure to accurately account for distortion artifacts in such scenarios greatly reduces the homography estimations and the subsequent placement of objects, lowering the quality of the overall user experience.

In this paper, we consider homography estimation using affine features of radially distorted images. Our core contributions consist of the following:
\begin{itemize}
    \item Theoretical justification and novel geometric constraints for one-sided radially distorted homographies from affine-covariant features and orientation-covariant features,
    \item Fast and accurate geometric solvers utilizing these new constraints, which are capable of using minimal and non-minimal samples, making it suitable for integration in modern robust estimation frameworks,
    \item Comparison to the state-of-the-art methods on synthetic and real data, where the proposed solvers are shown to be faster than traditional point-based solvers while maintaining the same level of accuracy.
\end{itemize}


\section{Related work}
A homography~$\HH$, also known as projective transformation, between two point correspondences~$\x_1\leftrightarrow\x_2$, where
$\x_1=[u_1,\,v_1,\,1]^{\T}$ and~$\x_2=[u_2,\,v_2,\,1]^{\T}$, respectively, obeys the relationship~$\x_2\sim {\HH} \x_1$,
where $\sim$ denotes equality up to scale.
In order to estimate a homography in the general case, one requires at least four point correspondences---two  constraints are
given for each point correspondences and there are in total eight degrees of freedom, since the homography is only determined up to scale. A classic algorithm for estimating the homography is the direct linear transformation~(DLT)~\cite{hartley-zisserman}. If additional information is known, the number of degrees of freedom may be reduced, and the geometric
information encoded into the governing equations, thus reducing the number of necessary image correspondences. This can be
achieved by mechanically constraining the camera
\cite{liang-pears,wadenback2013,wadenback2014,wadenback2016,valtonenoernhag2019,valtonenoernhag2019b,valtonen-ornhag-icpram-2020}
which can be extended to the two cameras connected by a rigid body motion~\cite{valtonenoernhag2018}.
If known, a common reference direction can be used, which can be either obtained by mechanical alignment or by utilizing inertial
measurement units (IMUs) to obtain the gravity
direction~\cite{saurer-etal-2017,ding-etal-2019-iccv,valtonen-ornhag-etal-icpr-2021}.
Affine features add additional knowledge of the scene, and have been used directly~\cite{barath-hajder-2017} or approximated 
from scale- and orientation-covariant features~\cite{ding-etal-cvpr-2024}, where the latter require only a single feature in
order to estimate the homography.

Fitzgibbon~\cite{fitzgibbon2001} introduced the \emph{division model} and developed a solver for simultaneous estimation of
a single distortion parameter and homography. This model has been shown very effective, as a single parameter is capable of
correcting a vast number of distortion profiles, and has paved way for a multitude of solvers
including~\cite{kukelova2015,valtonen-ornhag-etal-wacv-2021,ding-etal-2021-iccv,nakano-icpr-2025,wadenback-etal-2025}. Given distorted image
coordinates~$(u_d,\,v_d)$, the undistorted counterpart $(u,\,v)$ is represented in homogeneous coordinates by
\begin{equation}
    \begin{bmatrix}
        u_d \\ v_d \\ 1
    \end{bmatrix}
    \sim
    \begin{bmatrix}
        u \\ v \\ 1 + \lambda (u^2 + v^2)
    \end{bmatrix},
\end{equation}
where~$\lambda$ is a distortion parameter.

Despite its recent popularity, affine features have not been used extensively in estimating lens distortion. One reason 
for this might be that distortion affects the entire image, thus acting globally, and the local nature of affine features
are thought to not encoded such phenomena accurately. 
The work by Pritts~\etal{}~\cite{pritts-etal-cvpr-2018,pritts-etal-accv-2018} stands alone in addressing this subject. They proposed a set of solvers to
jointly undistort and affinely-rectified an image based on translation directions of coplanar repeated scene texture.
This is done by utilizing that an affine-rectifying homography transforms the image of the vanishing
line of the scene plane to the line at infinity. While demonstrating the feasibility of using affine-covariant features to estimate
lens distortions, the method does not estimate the homography between two images and makes specific assumptions on the scene 
geometry by requiring repeated patterns.

\section{Theory}
An affine correspondence (AC) is a triplet~$(\x_1,\x_2,\A)$ containing the point correspondences~$\x_1\leftrightarrow\x_2$, and a local affine transformation~$\A$. We use the definition introduced by~\cite{molnar-chetverikov-2014}, later also used by
\cite{barath-kukelova-2019-iccv,bentolila-francos-2014}, where~$\A$ is the first-order Taylor approximation of the projection
from 3D to 2D. Without any distortion artifacts, assuming there is a homography~$\HH$ such that~$\x_2\sim {\HH} \x_1$, and ordering the elements of $\mat{A}$ in row-major order,
\begin{equation}
    \begin{aligned}
        a_1 &= \frac{\partial u_2}{\partial u_1} = \frac{h_1-h_7u_2}{s} &\qquad&
        a_2 &= \frac{\partial u_2}{\partial v_1} = \frac{h_2-h_8u_2}{s} \\
        a_3 &= \frac{\partial v_2}{\partial u_1} = \frac{h_4-h_7v_2}{s} &\qquad&
        a_4 &= \frac{\partial v_2}{\partial v_1} = \frac{h_5-h_8v_2}{s}
    \end {aligned}
\end{equation}
where~$s=u_1h_7+v_1h_8+h_9$ is the projective depth.

\subsection{Affine constraints}
Now, consider the one-sided radial distortion homography
\begin{equation}\label{eq:1}
    \begin{bmatrix}
        u_2 \\ v_2 \\ 1
    \end{bmatrix}
    \sim\HH
    \begin{bmatrix}
        u_1 \\ v_1 \\ 1 + \lambda r_1^2
    \end{bmatrix},
\end{equation}
where $r_1^2 = u_1^2 + v_1^2$. Note that the selected side of the distorted image does not matter, as we may instead consider the inverse estimation of~$\HH$, \ie{}
\begin{equation}
\begin{bmatrix}
        u_1 \\ v_1 \\ 1 + \lambda r_1^2
    \end{bmatrix}
    \sim
    \HH^{-1}
    \begin{bmatrix}
        u_2 \\ v_2 \\ 1
    \end{bmatrix},
\end{equation}
hence we do not impose any restrictions using this parameterization.
Differentiation of~\eqref{eq:1} now yields 
\begin{align}\label{eq:mod_a1}
    a_1 &= \frac{\partial u_2}{\partial u_1} = \frac{h_1 + 2u_1\lambda h_3 - (h_7 + 2u_1\lambda h_9)u_2}{\sigma}, \\
    a_2 &= \frac{\partial u_2}{\partial v_1} = \frac{h_2 + 2v_1\lambda h_3 - (h_8 + 2v_1\lambda h_9)u_2}{\sigma}, \\
    a_3 &= \frac{\partial v_2}{\partial u_1} = \frac{h_4 + 2u_1\lambda h_6 - (h_7 + 2u_1\lambda h_9)v_2}{\sigma}, \\\label{eq:mod_a4}
    a_4 &= \frac{\partial v_2}{\partial v_1} = \frac{h_5 + 2v_1\lambda h_6 - (h_8 + 2v_1\lambda h_9 )v_2}{\sigma},
\end {align}
where $\sigma = u_1h_7 + v_1h_8 + (1+\lambda r_1^2)h_9$  is the radially corrected projective depth. Restructuring these
equations we get the following constraints
\begin{align}\label{eq:a1}
h_1 - (u_2+a_1u_1)h_7 - a_1v_1h_8 - a_1h_9 + 2u_1\lambda h_3 - (2u_1u_2+a_1r_1^2)\lambda h_9 &= 0, \\
h_2 - (u_2+a_2v_1)h_8 - a_2u_1h_7 - a_2h_9 + 2v_1\lambda h_3 - (2v_1u_2+a_2r_1^2)\lambda h_9 &= 0, \\
h_4 - (v_2+a_3u_1)h_7 - a_3v_1h_8 - a_3h_9 + 2u_1\lambda h_6 - (2u_1v_2+a_3r_1^2)\lambda h_9 &= 0, \\
h_5 - (v_2+a_4v_1)h_8 - a_4u_1h_7 - a_4h_9 + 2v_1\lambda h_6 - (2v_1v_2+a_4r_1^2)\lambda h_9 &= 0\;.\label{eq:a4}
\end{align}

\subsection{Orientation-covariant constraints}
Scale- and orientation-covariant features include a scale~$q_i$ and orientation~$\alpha_i$ which
are related to the elements of~$\mat{A}$.
In~\cite{barath-kukelova-2019-iccv}, two polynomials relating the affine features
to the corresponding scale- and orientation-covariant features were derived. They
are given by
\begin{align}
        q_1a_2a_3 - q_1a_1a_4 + q_2 &= 0, \label{eq:sift1}\\
        c_1s_2a_1 + s_1s_2a_2 - c_1c_2a_3 - c_2s_1a_4 &= 0, \label{eq:sift2}
\end{align}
where $s_i = \sin{\alpha_i}$ and~$c_i = \cos{\alpha_i}$.
We cannot directly substitute the constraints~\eqref{eq:sift1}--\eqref{eq:sift2} into~\eqref{eq:a1}--\eqref{eq:a4}; however,
we may study the ideal~$J$ generated by these six equations, and compute the elimination ideal to uncover the intrinsic
relationship between the scale- and orientation-covariant features and the one-sided radial homography. To do so we
use Macaulay2~\cite{macaulay2} to find the generators of $J_1=J\cap\C[h_1,\ldots,h_9,u_1,v_1,u_2,v_2,q_1,q_2,s_1,c_1,s_2,c_2,\lambda]$,
which are given by two polynomials, one of which is not dependent on the scale:
\begin{equation}\label{eq:ori}
\begin{aligned}
    &-h_1s_2c_1 - h_2s_1s_2 + h_4c_1c_2 + h_5s_1c_2 + h_7u_2s_2c_1 - h_7v_2c_1c_2 + h_8u_2s_1s_2\\
    &\qquad -h_8v_2s_1c_2 -2h_3u_1s_2c_1\lambda - 2h_3v_1s_1s_2\lambda + 2h_6u_1c_1c_2\lambda + 2h_6v_1s_1c_2\lambda \\
    &\qquad + 2h_9u_1u_2s_2c_1\lambda - 2h_9u_1v_2c_1c_2\lambda + 2h_9v_1u_2s_1s_2\lambda - 2h_9v_1v_2s_1c_2\lambda = 0
\end{aligned}
\end{equation}
While the other scale-dependent constraint could be used, one would still require three correspondences. Furthermore, it is of second order in the
elements of the homography. For these reasons, we do not pursue this further.

Note that in~\cite{barath-kukelova-2022-eccv}, another set of constraints were
derived from further geometric observations; however, when applied in this context, they do not contribute to any constraints other than~\eqref{eq:ori}.

\section{Solvers}
We may use the constraints~\eqref{eq:a1}--\eqref{eq:a4} or~\eqref{eq:ori} together with the point-based constraints
\begin{align}\label{eq:point1}
h_3 + h_1u_1 - h_9u_2 + h_2v_1 + h_3r_1^2\lambda - h_7u_1u_2 - h_8u_2v_1 - h_9r_1^2u_2\lambda &= 0, \\\label{eq:point2}
h_9v_2 - h_4u_1 - h_5v_1 - h_6 - h_6r_1^2\lambda + h_7u_1v_2 + h_8v_1v_2 + h_9r_1^2v_2\lambda &= 0,
\end{align}
which yields six equations per affine correspondence and three equations per orientation-covariant correspondences, respectively.
We adapt the approach used by Nakano~\cite{nakano-icpr-2025} as it has been shown superior to~\cite{fitzgibbon2001}; therefore,
let us introduce the lifting matrix
\begin{equation}
    \mat{Q}=\begin{bmatrix}
        0 & 1 & 0 & 0 \\
        0 & 0 & 1 & 0 \\
        \lambda & 0 & 0 & 1
    \end{bmatrix},
\end{equation}
and let $\mat{G}=\mat{H}\mat{Q}$. We may now write~\eqref{eq:point1}--\eqref{eq:point2} and \eqref{eq:a1}--\eqref{eq:a4} together as
\begin{equation}
    \mat{M}\vec{g}=\vec{0},
\end{equation}
where $\vec{g}=\vectorize(\mat{G})$, or, explicitly,
\setcounter{MaxMatrixCols}{12}
\begin{equation}
    \vec{g}=
    \begin{bmatrix}
        \lambda h_3 & h_1 & h_2 & h_3 & \lambda h_6 & h_4 & h_5 & h_6 & \lambda h_9 & h_7 & h_8 & h_9
    \end{bmatrix}^{\T},
\end{equation}
and
\begin{equation}
    \mat{M}=
    \resizebox{0.85\textwidth}{!}{$\displaystyle
    \begin{bmatrix}
        r_1^2 & u_1 & v_1 & 1 & 0 & 0 & 0 & 0 & -u_2r_1^2 & -u_2u_1 & -u_2v_1 & -u_2 \\
        0 & 0 & 0 & 0 & -r_1^2 &-u_1 &-v_1 &-1 & r_1^2v_2 & u_1v_2 & v_1v_2 & v_2 \\
        2u_1 & 1 & 0 & 0 & 0 & 0 & 0 & 0 & -(2u_1u_2+a_1r_1^2) & -(u_2+a_1u_1) & -a_1v_1 & -a_1 \\
        2v_1 & 0 & 1 & 0 & 0 & 0 & 0 & 0 & -(2v_1u_2+a_2r_1^2) & -a_2u_1 & -(u_2+a_2v_1) & -a_2 \\
        0 & 0 & 0 & 0 & 2u_1 & 1 & 0 & 0 & -(2u_1v_2+a_3r_1^2) & -(v_2+a_3u_1) & -a_3v_1 & -a_3 \\
        0 & 0 & 0 & 0 & 2v_1 & 0 & 1 & 0 & -(2v_1v_2+a_4r_1^2) & -a_4u_1 & -(v_2+a_4v_1) & -a _4
    \end{bmatrix},
    $}
\end{equation}
for the case of affine correspondences. Similarly, an expression for the orientation-covariant constraints~\eqref{eq:ori} is derived
\begin{equation}
    \mat{M}=
    \resizebox{0.65\textwidth}{!}{$\displaystyle
    \begin{bmatrix}
        r_1^2 & u_1 & v_1 & 1 & 0 & 0 & 0 & 0 & -u_2r_1^2 & -u_2u_1 & -u_2v_1 & -u_2 \\
        0 & 0 & 0 & 0 & -r_1^2 &-u_1 &-v_1 &-1 & r_1^2v_2 & u_1v_2 & v_1v_2 & v_2 \\
        m_{3,1} & m_{3,2} & m_{3,3} & 0 & m_{3,5} & m_{3,6} & m_{3,7} & 0 & m_{3,9} & m_{3,10} & m_{3,11} & 0
    \end{bmatrix},
    $}
\end{equation}
where
\begin{align}
    m_{3,1} &= -2(u_1s_2c_1 + v_1s_1s_2), \\
    m_{3,2} &= -s_2c_1, \\
    m_{3,3} &= -s_1s_2, \\
    m_{3,5} &= 2(u_1c_1c_2 + v_1s_1c_2), \\
    m_{3,6} &= c_1c_2, \\
    m_{3,7} &= s_1c_2, \\
    m_{3,9} &= 2u_1(u_2s_2c_1 - v_2c_1c_2) + 2v_1(u_2s_1s_2 - v_2s_1c_2), \\
    m_{3,10} &= u_2s_2c_1 - v_2c_1c_2, \\
    m_{3,11} &= u_2s_1s_2 - v_2s_1c_2\;.
\end{align}

\subsection{Minimal solution}
Given 1.5 ACs a minimal solver can be constructed. The approach is based on that of Nakano~\cite{nakano-icpr-2025} and only the coefficients of the matrix~$\mat{M}$ differ. We reproduce an abridged version for completeness, but
refer to~\cite{nakano-icpr-2025} for details.
The matrix $\mat{M}\in\R^{9\times 12}$ has a three-dimensional nullspace, hence can be expressed as a linear combination of its nullspace vectors
\begin{equation}\label{eq:alpha}
    \vec{g} = \alpha_1\vec{n}_1 + \alpha_2\vec{n}_2 + \alpha_3\vec{n}_3\;.
\end{equation}
Since the elements of~$\vec{g}$ are linearly dependent, these can be exploited, leading to a generalized eigenvalue problem
\begin{equation}
    \mat{A}\vec{\alpha} = \lambda \mat{B}\vec{\alpha},
\end{equation}
which has at most three real solutions. Backsubstituting $\vec{\alpha}$ in~\eqref{eq:alpha} one obtains~$\vec{g}$ from which the homography~$\mat{H}$ can be created.

\subsection{Non-minimal solution}
For $\geq 2$ ACs or $\geq 4$ orientation-covariant features a nonminimal solver can be constructed by solving the corresponding
least squares problem
\begin{equation}
    \min_{\norm{\vec{g}}=1}\norm{\mat{M}\vec{g}}^2,
\end{equation}
which is obtained by finding the smallest singular vector of~$\mat{M}\in\R^{6N\times 12}$, for $N$ affine correspondences, and~$\mat{M}\in\R^{3N\times 12}$, for $N$ orientation-covariant correspondences, respectively. The distortion coefficient can then be obtained by
\begin{equation}
    \lambda = \frac{\vec{g}_1^{\T}\vec{g}_4}{\norm{\vec{g}_4}^2},
\end{equation}
where~$\vec{g}_i$ is the $i$:th column of~$\mat{G}$. This is due to the construction of~$\mat{G}$, from which it follows that~$\vec{g}_1=\lambda\vec{g}_4$.

\begin{figure*}[t]
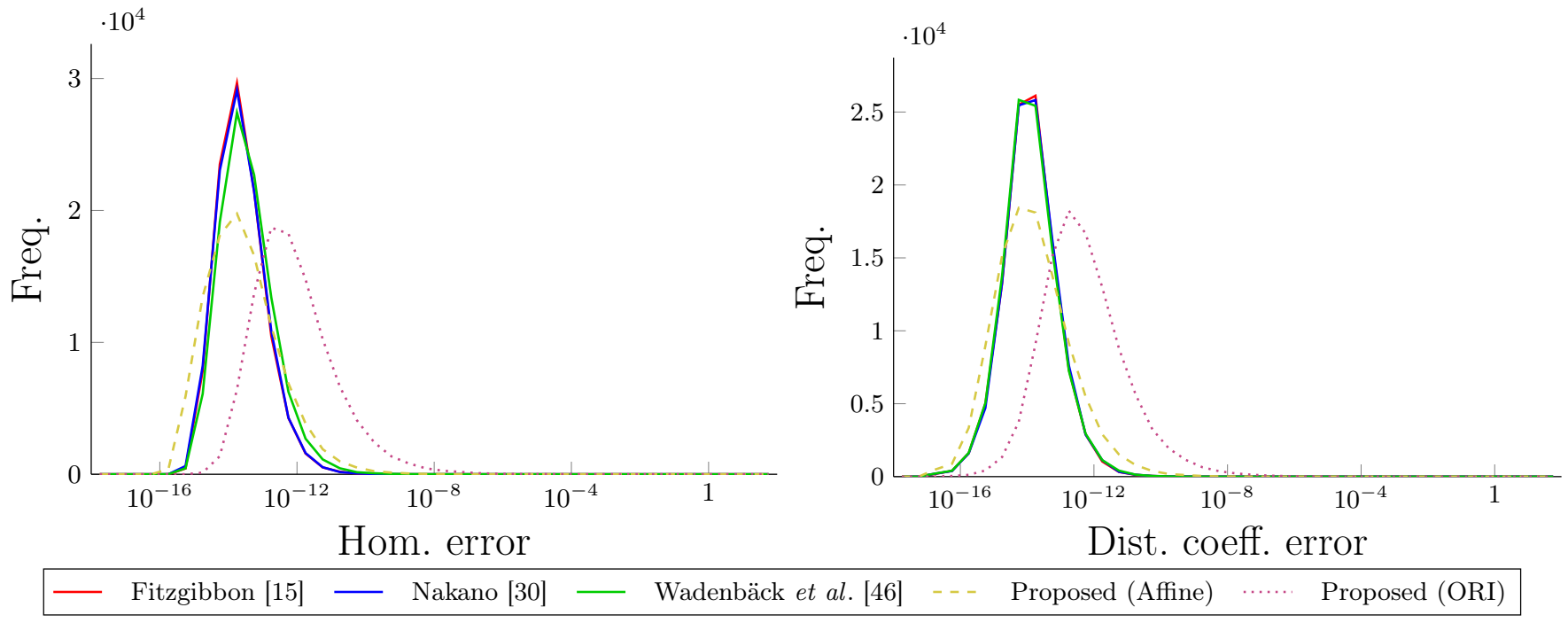

    \centering
    \includestandalone[width=0.48\linewidth]{plots/numerical_stability/numerical_stability_hom}%
    \includestandalone[width=0.48\linewidth]{plots/numerical_stability/numerical_stability_dist}\\%
    \includestandalone[width=0.91\linewidth]{plots/legend}%
    \caption{\emph{Numerical stability}. Homography errors (left) and distortion coefficient (right) for \num{100000} noise-free randomly generated problem instances.}
    \label{fig:stability}
\end{figure*}

\section{Experiments}
We compare our proposed solvers to the classic solver by Fitzgibbon~\cite{fitzgibbon2001}, and two recent ones by Nakano~\cite{nakano-icpr-2025} and Wadenbäck~\etal{}~\cite{wadenback-etal-2025}. These all consider the one-sided homography case and rely on point-based correspondences, for which they need 5 (or 4.5) correspondences in the minimal case. Note that in~\cite{fitzgibbon2001}, only the two-sided case was treated; however, in~\cite{nakano-icpr-2025} it was shown how to apply it to the one-sided case.
All solvers are implemented in C++ using Eigen. The results and timings are done on a standard laptop equipped with a 11th Gen Intel(R) Core(TM) i5-1145G7 @2.60GHz CPU. Since our method adapts the approach from~\cite{nakano-icpr-2025}, modifying only the coefficients of a matrix in the governing equations, the computational times remain comparable.

In all experiments, the homography error is defined as $\HH_{\textup{err}} \coloneqq \norm{\HH_{\textup{gt}}-\HH_{\textup{est}}}_{\textup{F}}$, where both $\HH_{\textup{gt}}$ and $\HH_{\textup{est}}$ are normalized to have unit Frobenius norm and positive determinant. Furthermore, we use the relative distortion coefficient error, defined as $\lambda_{\textup{err}} \coloneqq |\lambda_{\textup{gt}} - \lambda_{\textup{est}}| / |\lambda_{\textup{gt}}|$.

\subsection{Synthetic data}
\subsubsection{Numerical stability}
In order to test the numerical stability,  we generate a synthetic test scene
consisting of two calibrated cameras $\mat{P}_1 = [\mat{I}\;0]$ and $\mat{P}_1 = [\mat{R}\;\vec{t}]$.
Random 3D points $\mat{X}_i$ at a fixed random depth within $[0.1, 10]$ are generated, such that they are visible in both cameras assuming a 70{\degree}~field-of-view.
The length of the translation vector~$\vec{t}$ is set to 1 unit, and a random plane normal $\vec{n}$ is generated, which gives the ground truth homography~$\mat{H}\sim\gamma\mat{R}+\vec{t}\vec{n}^{\T}$, where the
plane equation satisfies~$\vec{n}^{\T}\vec{X}_i = \gamma$ for any point~$\vec{X}_i$ belonging to it.
For each~$\vec{X}_i$, an affine transformation associated
with it is generated using~\eqref{eq:mod_a1}--\eqref{eq:mod_a4}. This allows one to add noise to the affine correspondences in a controlled and interpretable manner, as discussed in~\cite{barath-kukelova-2022-eccv}. In~\cite{Ventura_2024_CVPR}, a method for converting affine features to scale- and orientation-covariant features was presented, which we apply
to generate synthetic orientation-covariant features (the scale is simply ignored).

A distortion coefficient in the range $[-0.2, -0.01]$ was generated, and the image correspondences distorted accordingly. This captures a low to medium level of distortion. The corresponding results are shown in~\cref{fig:stability}. All methods are numerically stable, and although the proposed solvers are slightly more susceptible to noise, this rarely impact the solvers in real scenarios, where execution time is generally more important.

\subsubsection{Noise sensitivity}
\begin{figure*}[t]
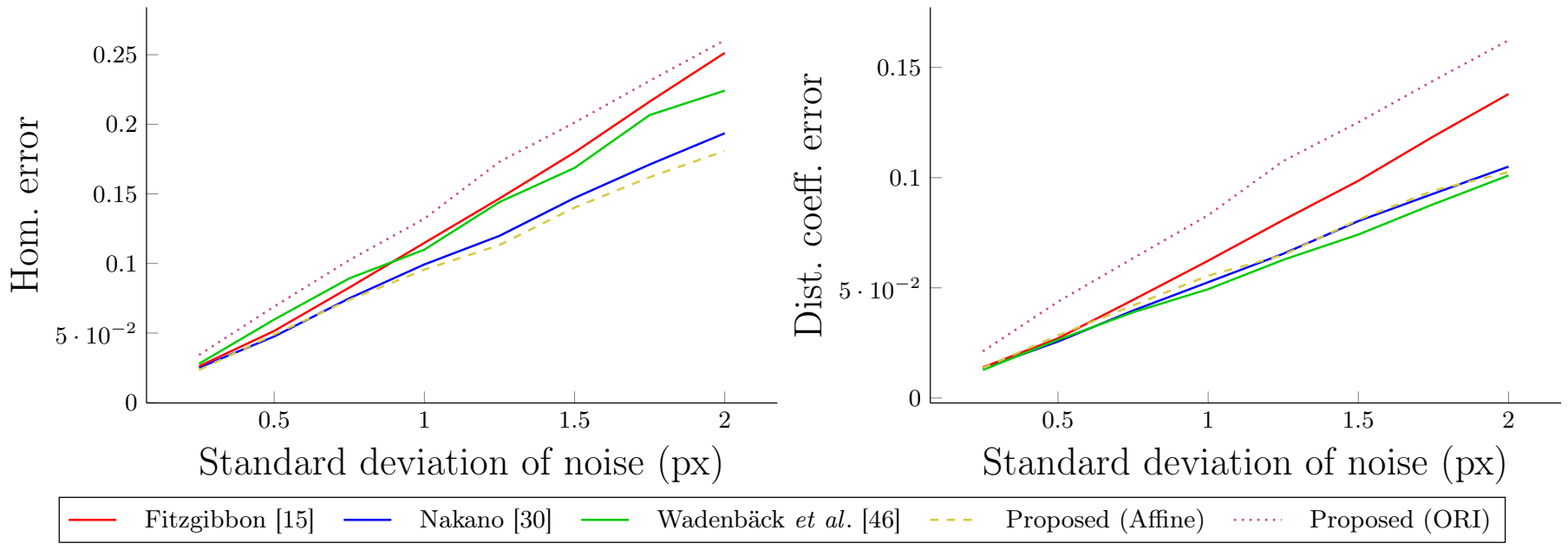

    \centering
    \includestandalone[width=0.49\linewidth]{plots/noise/noise_hom}%
    \includestandalone[width=0.49\linewidth]{plots/noise/noise_dist}\\
    \includestandalone[width=0.91\linewidth]{plots/legend}%
    \caption{%
        \emph{Sensitivity to noise}.
        Homography errors (left) and distortion coefficient (right) for different noise levels.
        Each point reports the median error of \num{10000} randomly generated problem instances (per noise level).
    }
    \label{fig:noise}
\end{figure*}

We further examine the robustness of the proposed solvers to Gaussian noise. In a similar synthetic setup, zero-mean Gaussian noise of varying standard deviation is introduced to the normalized point correspondences. In~\cref{fig:noise}, the median solver error over \num{10000} random instances per noise level is shown. We note that the orientation-covariant solver performs worst on both error metrics. This discrepancy may be attributed to the higher complexity inherent to the orientation-covariant constraint~\eqref{eq:ori}. In contrast, the affine solver achieves the best homography accuracy and performs comparably to state-of-the-art methods on distortion estimation.

\subsubsection{RANSAC integration}
Arguably, the main benefit of using affine- or orientation-covariant features is due to efficiently lowering the minimally required sample size, thus
reducing the necessary number of RANSAC iterations required to sample an all-inlier subset of the putatively matched image correspondences.
To test this, we proceed by generating \num{500} point correspondences, assuming a focal length of $f=1000$, and corresponding affine- and orientation-covariant features, per problem instance.
Furthermore, we add zero-mean Gaussian
image point noise with a standard deviation of~$\sigma=0.5$ pixels and corrupt the affine features and orientation-covariant features, equivalently.
An inlier threshold of \qty{5}{pixels} is used.
Finally, we inject outliers among the correspondences of varying ratios. The solvers are integrated in a standard RANSAC framework and the experiment is averaged over \num{1000} test instances for each outlier ratio.

\begin{figure*}[t!]
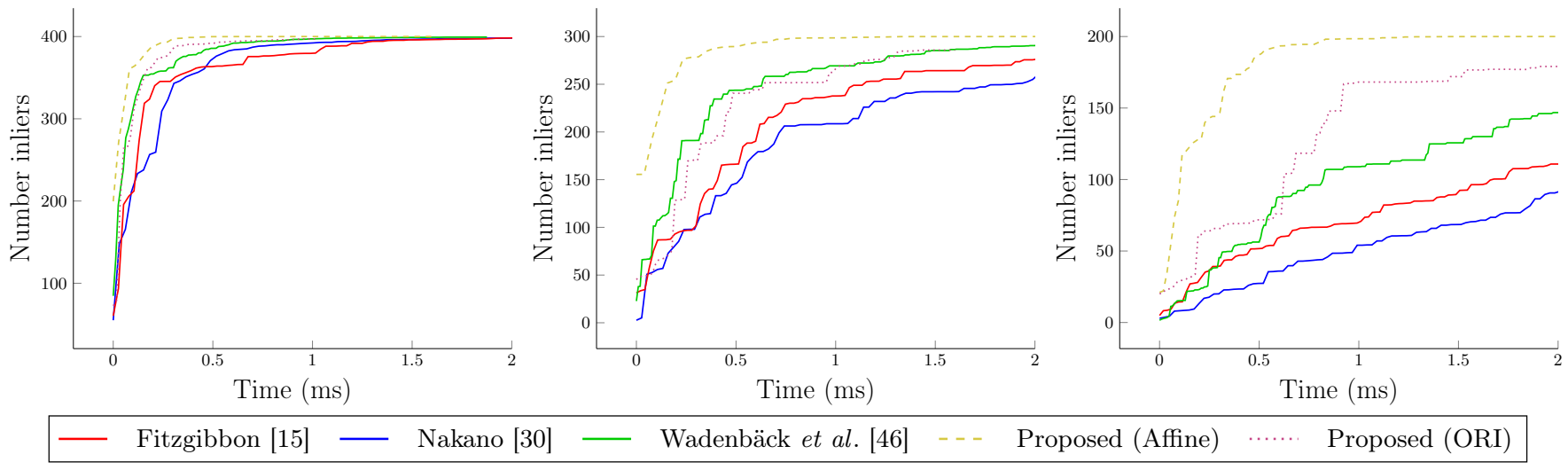

    \centering
    \includestandalone[width=0.32\linewidth]{plots/ransac/ransac_100}%
    \includestandalone[width=0.32\linewidth]{plots/ransac/ransac_200}%
    \includestandalone[width=0.32\linewidth]{plots/ransac/ransac_300}\\
    \includestandalone[width=0.91\linewidth]{plots/legend}%
    \caption{%
        \emph{RANSAC integration}.
        The cumulative number of inliers versus time for all evaluated solvers, for \qty{20}{\percent} outliers (left), \qty{40}{\percent} outliers (center), and \qty{60}{\percent} outliers (right).
    }
    \label{fig:ransac}
\end{figure*}

The results are shown in~\cref{fig:ransac}.
We note the orientation-covariant solver converging slowly initially, which may again be due to the complexity of the constraint~\eqref{eq:ori} and the previously observed higher sensitivity to noise,
\cf{}~\cref{fig:stability}; however, the affine-based method outperforms the state-of-the-art in all cases.
This demonstrates a key advantage of requiring fewer correspondences, and the consequential benefit in terms of execution time compared to point-based methods.

\subsection{Real data experiments}

\begin{figure}[t]
    \centering
    \includegraphics[width=0.532\linewidth]{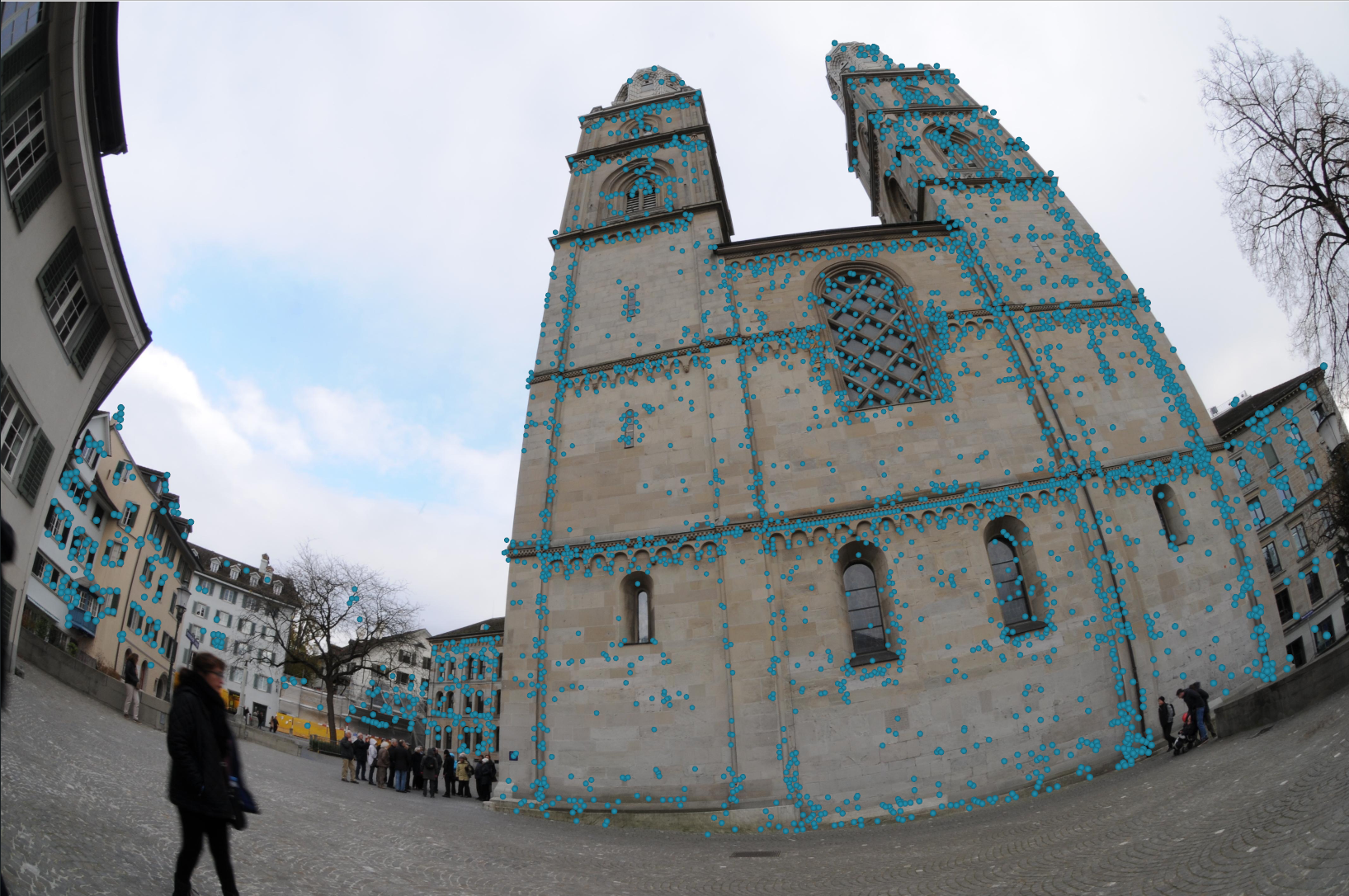}\quad
    \includegraphics[width=0.425\linewidth]{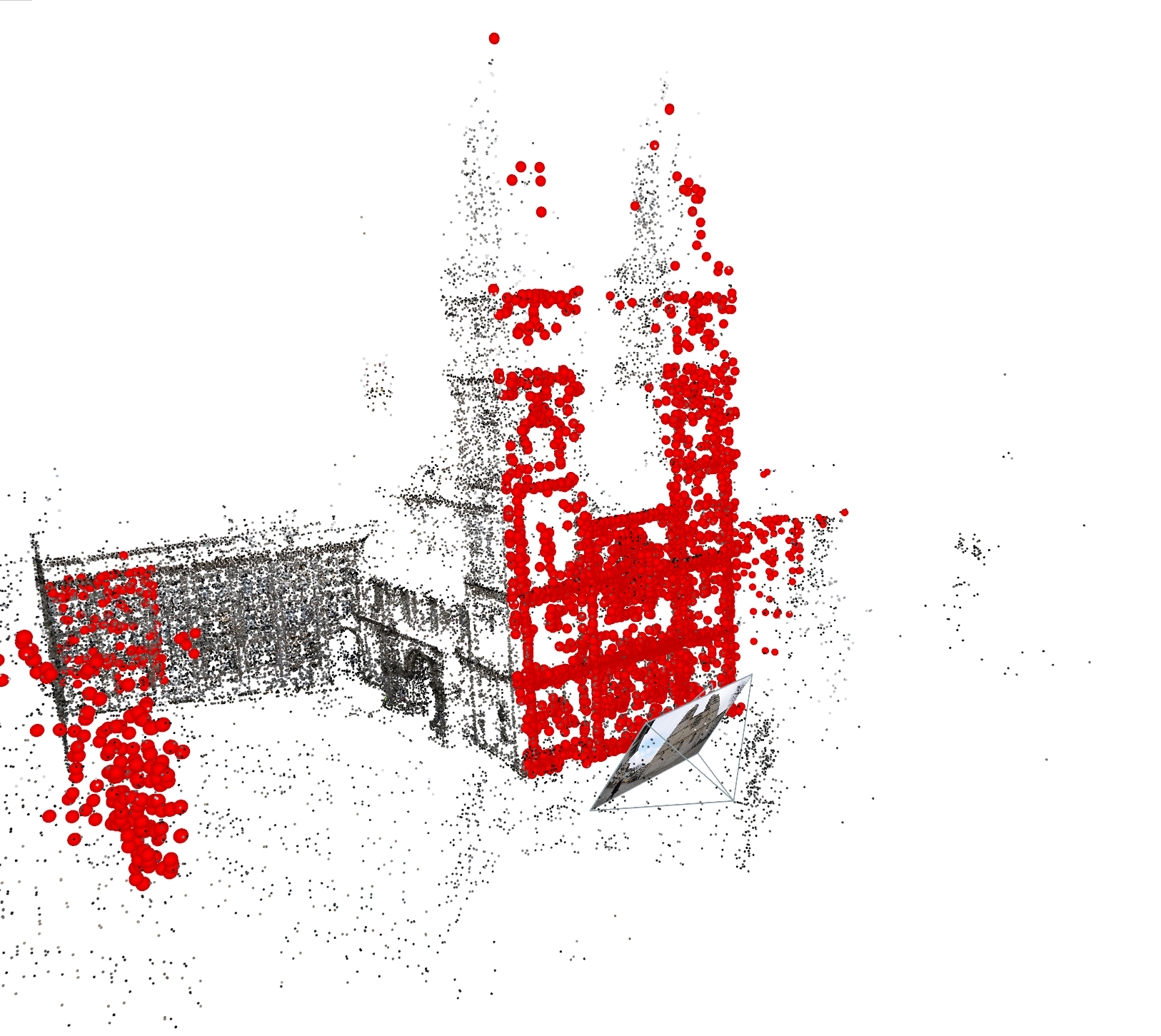}
    \caption{\emph{Grossmünster Church:} (\emph{Left:}) Location of extracted SIFT correspondences (blue) of a fisheye-distorted input image. (\emph{Right}:) Corresponding 3D points (red) showing the dominant plane including the two church towers. Note that the 3D model is not notably affected by distortion artifacts, as the rectification has been done during the reconstruction phase.}
    \label{fig:grossmunster}
\end{figure}

Due to the absence of real-life benchmarks for homography estimation with radial distortion, we use the Grossmünster Church dataset~\cite{larsson2020calibration}, as suggested in~\cite{wadenback-etal-2025}. This dataset provides real fisheye-distorted images where the distortion is not well modeled by a simple one-parameter division model, making it a challenging task.
The ground truth 3D reconstruction model is generated using RadialSfM~\cite{larsson2020calibration}, see~\cref{fig:grossmunster}. Following the setup of~\cite{wadenback-etal-2025},
we select images where a dominant plane is shown, namely the side with two church towers, and fit a plane based on the 3D reconstruction. The resulting dataset contains seven images, creating six consecutive image pairs to be used in the evaluation. 
The corresponding ground truth homographies are derived from the camera poses obtained from the reconstruction framework. The relative pose~$\bR=\bR_2\bR_1^{\T}$ and relative translation~$\bt=\bt_2-\bR\bt_1$ are used to compute $\bH_{\textup{gt}} = \bK(\bR-\bt(\bn')^{\T}/d)\bK^{-1}$, where $\bK$ is the shared intrinsic parameters across all views, $\bn'=\bR_1\bn$ is the relative plane normal, and~$d$ the corresponding depth.

Next, we employ the OpenCV implementation~\cite{opencv} of SIFT to extract image features for the point-based and orientation-covariant solvers. For the case of affine solvers,
AffNet~\cite{mishkin-etal-eccv-2018} was used, executed on top of the Difference of Gaussians detector implemented in OpenCV and HardNet descriptor~\cite{mishchuk2017working} from Kornia~\cite{eriba2019kornia}.
In all cases, Lowe's ratio test was carried out to establish putative matches with the default threshold set to 0.9.

As a robust estimator, we employed the LOMSAC~\cite{Lebeda2012BMVC}, which is configured to execute a minimum of \num{500} iterations, applying local optimization and final non-linear refinement.
Similarly to what was observed in~\cite{wadenback-etal-2025}, the average homography error is
almost identical for all solvers. The execution time differs, see~\cref{tab:grossmunster}, with a much preferable outcome for the affine- and orientation-covariant solvers proposed in this paper, 
demonstrating their applicability to challenging real-life problems.

Note that this setup is very similar to a mixed-reality setup which was earlier discussed, where the 3D environment is obtained using a calibrated camera (or, as in this case, the calibration profile is simultaneously obtained by the SLAM pipeline). As such, methods like this could likely be incorporated in future AR/MR devices.

\begin{table}[h!]
\centering
\caption{\emph{Grossmünster Church:} Results of the solvers in a LOMSAC framework. Best marked in \textbf{bold} and second best in \textit{italic}.}
\begin{tabular}{l|ccc|ccc|ccc}
\multirow{2}{*}{Method} & \multicolumn{3}{c|}{Hom. error} & \multicolumn{3}{c|}{Inlier ratio} & \multicolumn{3}{c}{Exec. time (ms)}  \\
                        & Median & Mean & Max             & Median & Mean & Max                           & Median & Mean & Max                  \\ 
\hline
Fitzgibbon~\cite{fitzgibbon2001}             & \textit{0.047} & \textbf{0.053} & \textbf{0.085} & 0.48          & 0.53          & 0.62          & 105.7          & 88.19          & 121.4         \\
Nakano~\cite{nakano-icpr-2025}               & 0.048          & \textbf{0.053} & \textbf{0.085} & 0.48          & 0.53          & 0.62          & 90.31          & 80.19          & 99.98         \\
Wadenbäck~\etal{}~\cite{wadenback-etal-2025} & 0.048          & \textbf{0.053} & \textbf{0.085} & 0.47          & 0.52          & 0.62          & 88.24          & 78.18          & 105.4         \\
Proposed (ORI)                               & \textit{0.047} & \textbf{0.053} & \textbf{0.085} & \textit{0.58} & \textit{0.59} & \textit{0.63} & \textit{60.70} & \textit{47.47} & \textit{61.52}\\
Proposed (Affine)                            & \textbf{0.046} & \textbf{0.053} & \textbf{0.085} & \textbf{0.73} & \textbf{0.73} & \textbf{0.78} & \textbf{20.39} & \textbf{22.89} & \textbf{35.84}         
\end{tabular}
\label{tab:grossmunster}
\end{table}

\begin{figure}[t]
    \centering
    \includegraphics[width=0.49\linewidth]{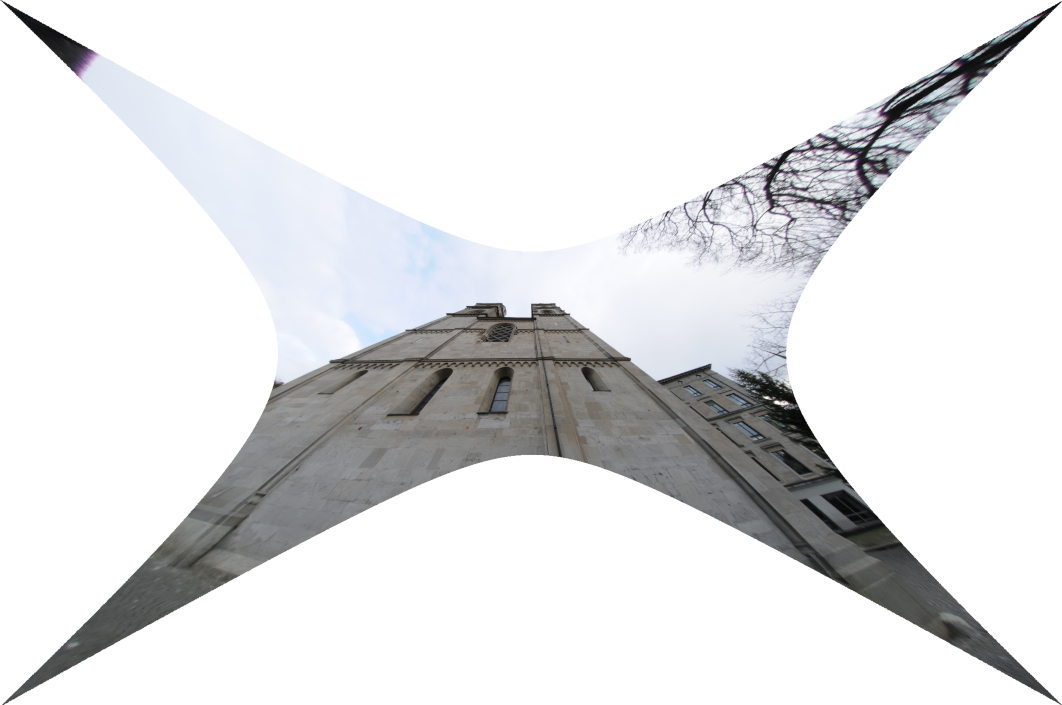}
    \includegraphics[width=0.49\linewidth]{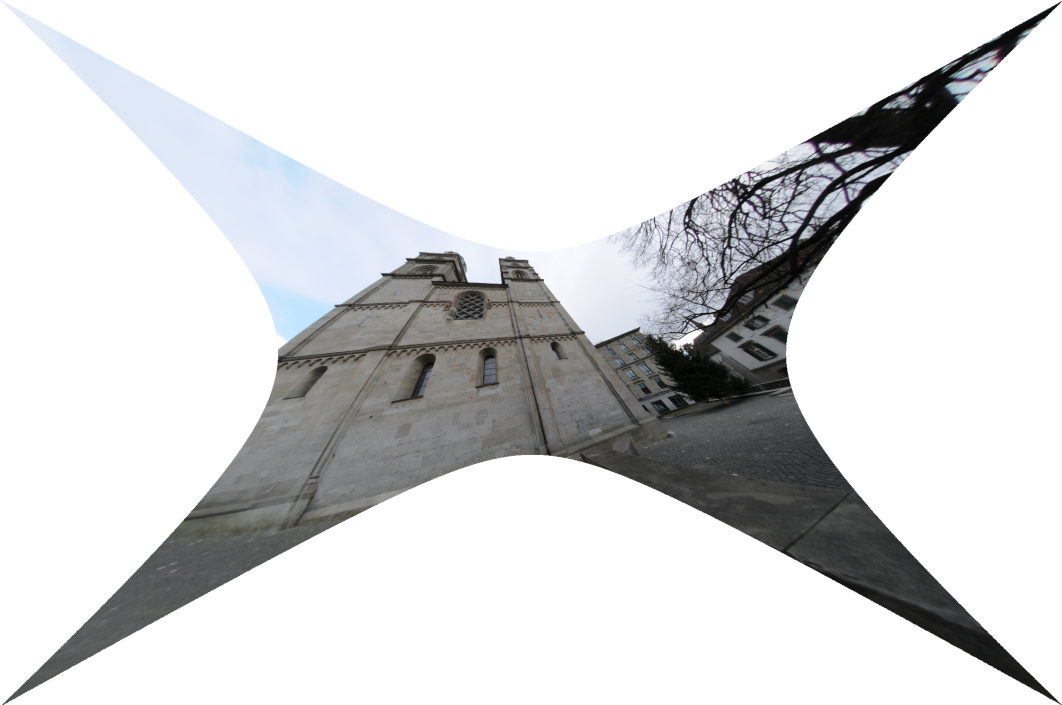}
    \caption{\emph{Grossmünster Church:} Rectified images using the estimated distortion parameter obtained by the proposed affine solver. Note that the edges of the church appear straight. This is a strong indication that the method performs well for real-life images, despite the severe fisheye distortion.}
    \label{fig:grossmunster-rad}
\end{figure}

As the fisheye camera distortion is not accurately modeled by the one-parameter division model, nor is used in the reconstruction framework, we cannot easily compute the error metric of the corresponding distortion coefficient; however, we may get some insights based on ocular inspection. For this reason, we provide two rectified example images from the dataset, where the affine method has been used to
estimate the distortion profile, see~\cref{fig:grossmunster-rad}. In both examples, physically straight lines remain geometrically faithful, with only minor artifacts appearing in the most distorted peripheral regions.

\section{Conclusions}
We have presented new minimal solvers for simultaneous homography and radial distortion estimation leveraging affine- or orientation-covariant features. This efficiently reduces the minimally required sample sizes compared to point-based methods, leading to significantly faster solvers of up to $5\times$ the state-of-the-art for point-based solvers. The affine-based solver demonstrates robust performance, achieving excellent homography accuracy and competitive distortion estimation. Its minimal sample size provides a decisive efficiency advantage in RANSAC, enabling faster and more reliable estimation in high-outlier scenarios. While the orientation-covariant solver is more noise-sensitive, it establishes the utility of this feature class for minimal solving. Real-world experiments on fisheye images confirm that both solvers match the accuracy of state-of-the-art methods while being significantly faster, proving their practical value for applications like AR/MR and photogrammetry.

%
%
\bibliographystyle{splncs04}
\bibliography{main}
\end{document}